\documentclass[
]{ceurart}

\usepackage{listings}
\usepackage{tabularx}
\usepackage{booktabs}

\usepackage[most]{tcolorbox}
\usepackage{xcolor}
\usepackage{caption}
\usepackage{listings}
\usepackage[most]{tcolorbox}

\newtcolorbox{promptbox}[1]{
  colback=gray!4,
  colframe=black!65,
  fonttitle=\bfseries\footnotesize,
  fontupper=\ttfamily\scriptsize,
  title=#1,
  breakable,
  boxrule=0.5pt,
  left=4pt, right=4pt, top=3pt, bottom=3pt,
  enhanced,
}

\newcommand{\promptcap}[1]{%
  \par\smallskip\noindent{\footnotesize #1}\par\smallskip
}

\begin{document}

\copyrightyear{2026}
\copyrightclause{Copyright for this paper by its authors.
  Use permitted under Creative Commons License Attribution 4.0
  International (CC BY 4.0).}

\conference{CLEF 2026 Working Notes, 21 -- 24 September 2026, Jena, Germany}

\title{From Voting to Agent Collaboration: Answer-Type-Aware LLM Pipelines for BioASQ 14b}

% \title[mode=sub]{<DMIS LAB> at CLEF 2026}

% \tnotetext[1]{You can use this document as the template for preparing your
%   publication. We recommend using the latest version of the ceurart style.}

\author[1]{Taeyun Roh}[%
orcid=0009-0009-7096-6410,
email=nrbsld@korea.ac.kr,
]
\fnmark[1]

\author[1]{Eunha Lee}[%
orcid=0009-0008-3165-6973,
email=eunhalee@korea.ac.kr,
]
\fnmark[1]

\author[2]{Wonjune Jang}[%
orcid=0009-0005-6216-3096,
email=dnjswnswkd03@mju.ac.kr,
]
\fnmark[1]

\author[1]{Sohyun Chung}[%
orcid=0009-0005-8354-666X,
email=sohyunjung@korea.ac.kr,
]
\fnmark[1]

\author[1]{Junha Jung}[%
orcid=0009-0003-1675-8414,
email=goodjungjun@korea.ac.kr,
]
\fnmark[1]

\author[1,3]{Jaewoo Kang}[%
email=kangj@korea.ac.kr,
]
\cormark[1]

\address[1]{Department of Computer Science and Engineering, Korea University, Seoul, 02841, Republic of Korea}
\address[2]{Department of Mathematics, Myongji University, Yongin, 17058, Republic of Korea}
\address[3]{AIGEN Sciences, Seoul, 04778, Republic of Korea}

\fntext[1]{These authors contributed equally to this work.}
\cortext[1]{Corresponding author.}

% \cortext[1]{Corresponding author.}
% \fntext[1]{These authors contributed equally.}

\begin{abstract}
  Biomedical question answering requires not only accurate extraction of information from scientific literature but also reliable integration of evidence across multiple documents. This study presents a question-type-specific large language model (LLM) framework for BioASQ 14b Task B, designed to improve answer robustness and evidence grounding in biomedical question answering. Rather than applying a single prompting strategy to all questions, the framework selects different inference procedures for yes/no, factoid, and list questions according to their distinct reasoning and evaluation requirements. For yes/no questions, snippet shuffling and self-reflection are used to reduce sensitivity to evidence ordering and improve decision stability. For factoid questions, full-snippet input is combined with chain-of-thought-based in-context learning to support accurate biomedical entity identification. For list questions, a multi-agent architecture is employed, in which evidence extraction, candidate generation, answer verification, and final aggregation are handled collaboratively. Preliminary experiments on BioASQ 13b were used to identify effective inference strategies for each question type, and the resulting framework was subsequently evaluated in the official BioASQ 14b Task B challenge. In the official evaluation, our framework showed competitive performance across multiple batches and achieved first place in the factoid subtask of Batch 4. These results demonstrate the effectiveness of combining question-type-specific inference, ensemble prediction, and agent-based verification for reliable biomedical question answering.

\end{abstract}


\maketitle

\section{Introduction}

Biomedical literature is expanding at a pace that makes it increasingly difficult for researchers and healthcare professionals to identify precise and reliable answers to specialized questions. Important evidence is often distributed across multiple articles, expressed using heterogeneous terminology, or reported in partially complementary and sometimes conflicting contexts. As a result, biomedical QA systems cannot rely on document retrieval alone. They must read evidence in context, connect findings scattered across snippets, and return answers that are concise enough for evaluation while still being grounded in the source literature. The BioASQ challenge has served as a major benchmark for advancing such capabilities by evaluating systems on biomedical semantic indexing and question answering tasks~\cite{Tsatsaronis2015BioASQ,nentidis2026bioasq14}. In particular, BioASQ Task B requires systems to generate exact and ideal answers from evidence snippets extracted from biomedical literature, making it a suitable testbed for evaluating evidence-grounded reasoning in large language models (LLMs)~\cite{nentidis2026bioasq14,Krithara2023BioASQQA}.

BioASQ Task B includes several answer formats, among which yes/no, factoid, and list questions pose distinct reasoning challenges. Yes/no questions require a stable binary decision based on potentially dispersed or conflicting evidence. Factoid questions demand accurate identification of a specific biomedical entity, such as a drug, gene, disease, or molecular target, where minor variations in terminology can directly affect exact-match evaluation. List questions are more challenging in terms of coverage and verification because systems must retrieve multiple correct entities while avoiding unsupported or redundant predictions. Although all three types are grounded in the same collection of snippets, they differ substantially in their reasoning requirements and evaluation criteria~\cite{Krithara2023BioASQQA}. This suggests that a single inference strategy may not be equally effective across all question types.

Recent LLMs have demonstrated strong capabilities in medical question answering and reasoning, particularly in retrieval-augmented settings where relevant contextual evidence is provided~\cite{lievin2024medical, singhal2025toward, zhang2025leveraging,jung2026breaking}. However, their application to biomedical QA remains constrained by several practical issues. First, model outputs can be sensitive to the order and composition of retrieved snippets, producing unstable predictions even when the underlying evidence is unchanged~\cite{kim2025prompting, liu2023lostmiddlelanguagemodels, zhang2025leveraging}. Second, when multiple snippets contain overlapping or conflicting information, an LLM may fail to identify which evidence should dominate the final answer~\cite{Zong_2025, lewis2021retrievalaugmentedgenerationknowledgeintensivenlp}. Third, entity-oriented questions require careful output normalization and evidence grounding, since plausible but unsupported biomedical terms may be generated~\cite{Merker2024MiBiAB, nentidis2025overviewbioasq2025thirteenth}. Finally, hallucinated answers are especially problematic in the biomedical domain, where reliability and traceability are essential~\cite{zhu-etal-2025-trust, asgari2025framework}. Prior work on retrieval-augmented biomedical LLMs has shown that domain-specific retrieval and self-reflection can improve answer reliability and evidence use~\cite{jeong2024improvingmedicalreasoningretrieval, ateia2024bioragentretrievalaugmentedgenerationshowcasing}. These limitations motivate inference procedures that explicitly account for evidence organization, answer verification, and question-specific characteristics.

In this work, we present a type-specific LLM-based ensemble and agentic QA framework developed for BioASQ 14b Task B. Rather than applying a uniform inference procedure across all questions, our framework assigns specialized strategies according to question type. The strategies were selected through preliminary experiments on BioASQ 13b data and subsequently deployed in the official BioASQ 14b evaluation. Previous studies have independently demonstrated the robustness of both LLM-based ensembles and agentic QA frameworks~\cite{panou2024farming,roh2026clag}, but our contribution lies in unifying these two approaches. For yes/no questions, we apply snippet shuffling together with self-reflection~\cite{shinn2023reflexionlanguageagentsverbal} to reduce sensitivity to evidence ordering and improve the consistency of binary decisions. For factoid questions, we combine full-snippet input with chain-of-thought-based in-context learning \cite{wei2023chainofthoughtpromptingelicitsreasoning, wang2023selfconsistencyimproveschainthought} to support precise identification of biomedical entities. For list questions, we employ a multi-agent collaboration architecture in which evidence extraction, candidate generation, verification, and final answer aggregation are performed in separate but coordinated stages~\cite{tang2024medagentslargelanguagemodels, ateia2024bioragentretrievalaugmentedgenerationshowcasing}. Similar agent-oriented architectures have recently been explored for biomedical question answering in BioASQ~\cite{angulo2025aqams}.

To further improve robustness, we apply the selected type-specific strategies to multiple LLMs and integrate their outputs using task-appropriate ensemble procedures. This design aims to mitigate model-specific bias while preserving the distinct requirements of each answer format. In the official BioASQ 14b Task B evaluation, \texttt{ku\_dmis} demonstrated competitive performance across the evaluated batches and question types. Notably, it ranked first in the factoid subtask of Batch 4 according to MRR~\cite{BioASQ14bResults}. These results support the effectiveness of combining question-type-specific inference, ensemble prediction, and explicit verification mechanisms for reliable LLM-based biomedical question answering.

\section{Task setting}

The BioASQ initiative aims to advance research on automated systems for answering biomedical questions. BioASQ 14b uses English biomedical questions and benchmark data prepared by biomedical experts. In Phase B of Task 14b, the organizer provides the same questions released in Phase A, together with gold relevant articles and snippets. Systems may return an ideal paragraph answer for any question type and exact answers for \textbf{yes/no, factoid}, and \textbf{list} questions \cite{BioASQ14bGuidelines}. The 2026 test set was released in four batches, with Phase B answers due after gold snippets became available for each batch \cite{BioASQ14bGuidelines}. \\

% For exact answers, the official result page reports separate metrics for each type. Yes/no questions are evaluated with accuracy, F1 for yes, F1 for no, and macro F1. Factoid questions are evaluated with strict accuracy, lenient accuracy, and mean reciprocal rank (MRR). List questions are evaluated with mean precision, recall, and F-measure \cite{BioASQ14bResults}. We submitted only exact-answer outputs under system identifiers of the form \texttt{ku\_dmis}, \texttt{ku\_dmis\_2}, and so on. The BioASQ guidelines allow up to five systems per participant group \cite{BioASQ14bGuidelines}; the result analysis in this paper follows the user's requested aggregation rule by taking, for each metric in each batch, the highest value among the available \texttt{ku\_dmis} submissions.

\noindent \textbf{Yes/No.} \hspace{1em}
Given the supplied snippets, the system must determine whether the appropriate response is "yes" or "no." A representative example is: \textit{"Can tumor cells be reprogrammed into immune cells?"} In many cases, arriving at a correct yes/no decision requires extracting and aggregating information from several snippets rather than depending on any single passage. Performance is measured using either \textbf{accuracy} or the \textbf{macro-averaged F1-score}, in which "yes" and "no" are handled as independent classes. 
For yes/no questions, we followed the official BioASQ evaluation protocol for exact answers introduced in BioASQ6 \cite{malakasiotis2020evaluation}. The F1-score is computed independently for the “yes” and “no” classes, yielding $F1_y$ and $F1_n$, respectively.  For $F1_y$, "yes" is treated as the positive class, whereas for $F1_n$, "no" is treated as the positive class. The official score is the macro-averaged F1 score:
\[
    maF1 = \frac{F1_y + F1_n}{2},
\]

\noindent which assigns equal importance to performance on both answer classes. \\

\noindent \textbf{Factoid.} \hspace{1em}
Questions in this category call for a concise response that conveys the specific piece of factual content being asked about. For instance, \textit{"What is the most common genotoxic factor that humans are exposed to?"} The system is required to return a list up to 5 entity names, ordered by decreasing confidence. Gold-standard responses are represented as a nested list, with each inner list grouping together synonymous or equivalent surface forms of the same correct answer. \\

Factoid questions are evaluated using \textbf{strict accuracy (SAcc), lenient accuracy (LAcc), and mean reciprocal rank (MRR)}. SAcc considers an answer correct only when the gold entity name or one of its synonyms appears as the first item in the system’s returned list, whereas LAcc considers it correct if the gold entity or synonym appears anywhere in the returned list. Although SAcc and LAcc are reported for completeness, the official metric for factoid exact answers is MRR \cite{malakasiotis2020evaluation}. MRR evaluates the rank position of the first correct answer in the returned list and gives higher scores when the gold answer or its synonym appears closer to the top.  In the definition below, for each factoid question $q_i$ the topmost position that contains the golden entity name is searched. If the topmost position is the $j$-th one, then $r(i)=j$.
\[
\mathrm{MRR} = \frac{1}{n} \cdot \sum_{i=1}^{n} \frac{1}{r(i)},
\]

\noindent where $n$ is the number of factoid questions. \\

\noindent \textbf{List.} \hspace{1em}
List-type questions require systems to return multiple correct entity names. A representative example is: ``List signaling molecules (ligands) that interact with the receptor EGFR?'' Unlike factoid questions, which expect a single best answer or a small ranked list of aliases, list questions evaluate whether the system retrieves as many distinct gold-standard entities as possible. Each gold entity may include synonyms, and if the system returns a synonym of a gold entity, it is counted as a correct match. Duplicate mentions of the same entity are counted only once. For each list question, the system’s returned list is compared with the gold-standard list using precision, recall, and F1-score. The order of returned answers does not affect the score; only the overlap between the predicted and gold answer sets is considered. Scores are averaged across all list questions, and the official BioASQ evaluation measure for list-type questions is the \textbf{mean F1-score} \cite{malakasiotis2020evaluation}.

\section{System overview}

% \begin{figure}[t]
% \centering
% \setlength{\fboxsep}{6pt}
% \fbox{%
% \begin{minipage}{0.94\linewidth}
% \small
% \textbf{Input:} question, question type, gold BioASQ snippets\\[2pt]
% $\rightarrow$ \textbf{Type router}\\[2pt]
% \hspace*{1em}\textbf{Yes/No:} multi-LLM repeated inference + snippet shuffling + majority vote + selective verification\\
% \hspace*{1em}\textbf{Factoid:} full-snippet multi-LLM inference + CoT-based in-context examples + consensus filtering\\
% \hspace*{1em}\textbf{List:} evidence analyst + reasoning agent + verification agent + aggregation agent\\[2pt]
% $\rightarrow$ \textbf{Answer normalizer and BioASQ JSON formatter}
% \end{minipage}}
% \caption{High-level architecture of the KU-DMIS exact-answer system.}
% \label{fig:pipeline}
% \end{figure}

\begin{figure}[t]\centering\includegraphics[width=0.9\linewidth]{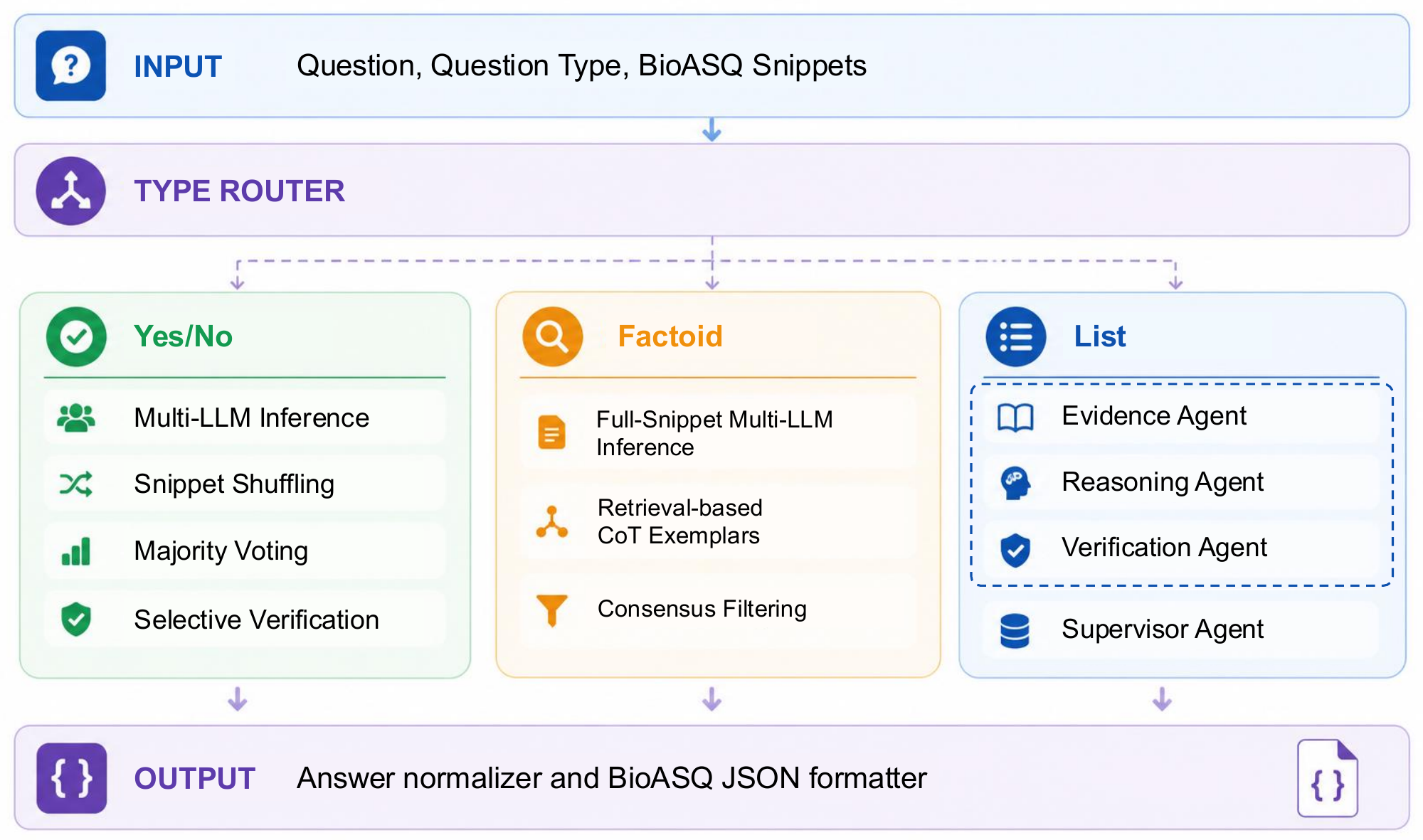}\caption{Overall architecture for exact-answer generation pipeline in BioASQ 14b.}\label{fig:pipeline}\end{figure}

Figure~\ref{fig:pipeline} summarizes the overall architecture. The system first routes each question according to its BioASQ type. Each branch receives the question body and the gold snippets. We did not use a single universal prompt; instead, each branch uses a prompt schema and aggregation strategy tailored to its expected answer space. The final stage validates the output format before submission, enforcing yes/no labels for binary questions, ranked lists of up to five entries for factoid questions, and list-style entity outputs for list questions.

The system is inference-time and prompt-driven. Multiple LLMs are used as complementary biomedical reasoners, and their outputs are aggregated to reduce sensitivity to a single model's prompt interpretation. The main aggregation logic differs by question type: majority voting for yes/no, consensus filtering for factoid answers, and agentic evidence verification for list answers.

\section{Methods}

During the BioASQ-14b challenge period, official test results for the submitted batches were not available. Therefore, we used the BioASQ-13b test set as a development benchmark to compare candidate prompting strategies, model backbones, snippet-processing methods, and agent-based variants. Tables~\ref{tab:13b_yesno}, \ref{tab:13b_factoid}, and \ref{tab:13b_list} report these preliminary experiments for yes/no, factoid, and list questions, respectively. Based on these development results, we selected the most robust strategy for each question type and applied the system to the official BioASQ-14b evaluation.

\subsection{Yes/No questions}

% The yes/no subsystem uses an ensemble of four LLM configurations: GPT-5 thinking, GPT-4o, GPT-5 pro, and Gemini 3.0 thinking. In earlier internal experiments, GPT-4o was competitive with, and sometimes stronger than, GPT-5-family models for yes/no cases, so it was retained in the final ensemble.

% For each question, we perform snippet shuffling to reduce dependence on snippet order. Each model is queried four times with different snippet orders, and each run is constrained to return one of \texttt{yes}, \texttt{no}, or \texttt{none}. The initial answer is obtained by majority vote over the sampled outputs. Allowing \texttt{none} during intermediate inference helps identify uncertain cases, but the final submitted answer must be a BioASQ-valid \texttt{yes} or \texttt{no}.

% A verification agent is applied selectively. If all model outputs are consistent, the majority decision is submitted directly. If at least one model output differs, the case is passed to an agentic verification pipeline. The first component, an evidence-structuring agent, groups the snippets into positive evidence, negative evidence, and uncertainty evidence with respect to the question. The second component, a final-decision agent, receives the structured evidence and all model outputs, then produces a final \texttt{yes}, \texttt{no}, or \texttt{none}. If \texttt{none} remains, the evidence is rechecked and the prompt is tightened to force a binary decision; this re-adjudication is allowed up to two times.

The yes/no subsystem is designed to produce stable binary decisions from conflicting evidence. As shown in Table~\ref{tab:13b_yesno}, several methods achieved strong performance on the BioASQ-13b development benchmark. The previous baseline denotes the system that achieved state-of-the-art performance in the BioASQ-13b yes/no subtask~\cite{kim2025prompting}. We extended this baseline with self-reflection, repeated inference, and a verification agent to improve the robustness of yes/no prediction.

Based on these findings, the final yes/no subsystem employs an ensemble of four LLM configurations: GPT-5 Thinking, GPT-5 Pro~\cite{openai_gpt5}, GPT-4o~\cite{openai_gpt4o} and Gemini 3.0 Thinking~\cite{google_gemini3}. GPT-4o was included because it remained competitive with GPT-5-family models and occasionally outperformed them in the BioASQ-13b development experiments.

For each question, snippets are shuffled to reduce sensitivity to snippet order. Each model is queried four times with different snippet orders, and each run is constrained to return a binary answer, either \texttt{yes} or \texttt{no}. The initial answer is then determined by majority vote over the sampled outputs. Because conflicting outputs indicate that the evidence may support more than one interpretation, such cases are further examined by a verification agent rather than being resolved solely by majority voting.

The verification agent is applied selectively. If all model outputs are consistent, the majority decision is submitted directly. If at least one model output differs, the case is passed to an agentic verification pipeline. The first component, an evidence-structuring agent, groups the snippets into positive evidence, negative evidence, and uncertainty evidence with respect to the question. The second component, a final-decision agent, receives the structured evidence and all model outputs, then produces a final \texttt{yes}, \texttt{no}, or \texttt{none}. If \texttt{none} remains, the evidence is rechecked and the prompt is tightened to force a binary decision; this re-adjudication is allowed up to two times.

\begin{table}[!htbp]
\caption{Evaluation on the BioASQ-13b test set for yes/no questions. The best score in each column is highlighted in bold. ICL2 denotes in-context learning with two demonstration examples.}
\label{tab:13b_yesno}
\centering
\scriptsize
\resizebox{0.8\linewidth}{!}{%
\begin{tabular}{llrrrr}
\toprule
Model & Method & Acc. & F1 Yes & F1 No & Macro F1 \\
\midrule

\multirow{7}{*}{GPT-4o}
& Full snippets & 0.9412 & \textbf{1.0000} & 0.8889 & 0.9444 \\
& Full snippets + ICL2 & 0.9412 & \textbf{1.0000} & 0.8889 & 0.9444 \\
& Verifier + Reviser + ICL2 & 0.9412 & \textbf{1.0000} & 0.8889 & 0.9444 \\
& Snippet highlight & 0.9412 & \textbf{1.0000} & 0.8889 & 0.9444 \\
& Snippet highlight + ICL2 & 0.9412 & \textbf{1.0000} & 0.8889 & 0.9444 \\
& Previous baseline & \textbf{1.0000} & \textbf{1.0000} & \textbf{1.0000} & \textbf{1.0000} \\
& Previous baseline + self-reflection & \textbf{1.0000} & \textbf{1.0000} & \textbf{1.0000} & \textbf{1.0000} \\

\midrule

\multirow{4}{*}{GPT-5.2}
& Full snippets & 0.9412 & 0.9600 & 0.8889 & 0.9244 \\
& Full snippets + CoT ICL2 & 0.9412 & 0.9600 & 0.8889 & 0.9244 \\
& Snippet filtering top-10 & 0.9412 & 0.9600 & 0.8889 & 0.9244 \\
& Agent & 0.9412 & 0.9600 & 0.8889 & 0.9244 \\

\midrule

\multirow{3}{*}{GPT-5.4}
& Full snippets & 0.9412 & 0.9600 & 0.8889 & 0.9244 \\
& Snippet filtering top-10 & 0.9412 & 0.9600 & 0.8889 & 0.9244 \\
& Agent & 0.9412 & 0.9600 & 0.8889 & 0.9244 \\

\midrule

\multirow{2}{*}{Claude}
& Sonnet-4.6 Agent & 0.9412 & 0.9600 & 0.8889 & 0.9244 \\
& Opus-4.6 Agent & 0.9412 & 0.9600 & 0.8889 & 0.9244 \\

\bottomrule
\end{tabular}}
\end{table}

\subsection{Factoid questions}

The factoid subsystem is designed to produce a ranked list of concise biomedical entities or phrases that answer a given question. Since BioASQ factoid evaluation is sensitive to exact answer forms, the subsystem must recover both the correct entity and its expected surface form. Table~\ref{tab:13b_factoid} shows that in-context learning and full-snippet input were effective in the BioASQ-13b development experiments. GPT-4o with ICL-based prompting achieved strong strict accuracy and MRR, while GPT-5-family models showed competitive lenient accuracy. These findings motivated a factoid pipeline that combines full-snippet reasoning, retrieved in-context examples, and multi-model consensus.

Because factoid prediction depends heavily on answer form, the demonstrations used for in-context learning must provide more than output-format guidance. They should also expose the model to similar question types, evidence patterns, and entity forms. We therefore use retrieval-based ICL, where semantically similar training questions are selected as demonstrations. This design follows prior findings that the choice of in-context examples can strongly affect downstream performance \cite{liu2021makesgoodincontextexamples}.

For each test question, we retrieve semantically similar training questions using BioBERT-based question embeddings \cite{lee2020biobert}. The retrieved examples are stored in a precomputed ICL index, and two examples are inserted into the prompt at inference time. This provides question-specific demonstrations that are closer to the target biomedical context than manually selected generic examples.

We further augment the retrieved demonstrations with verified chain-of-thought exemplars. Rather than encouraging verbose final outputs, these exemplars demonstrate the evidence-to-answer process required for factoid QA: identifying the answer-bearing snippet, excluding related but incorrect entities, and producing a concise canonical answer. For training examples referenced by the ICL index, we use GPT-5.2~\cite{openai_gpt52} to generate a short reasoning trace and an exact answer, retaining only examples whose generated answer matches the gold answer. When available, the verified exemplar replaces the corresponding plain ICL example during inference.

For each target question, all golden snippets are provided to the model to avoid missing answer mentions or disambiguating evidence. For ICL examples, we use at most three snippets per example to control prompt length.

The same prompting strategy is applied to a heterogeneous pool of LLMs, including Claude Opus 4.6~\cite{anthropic_claude_opus46}, Claude Sonnet 4.6~\cite{anthropic_claude_sonnet46}, GPT-5.2~\cite{openai_gpt52}, GPT-5.4~\cite{openai_gpt54}, Gemini 3.1 Pro Preview~\cite{google_gemini31pro}, and Gemini 3 Flash Preview~\cite{google_gemini3flash}. Each model outputs a JSON array of exact-answer candidates. After normalizing superficial formatting differences, candidates are aggregated by voting. Answers appearing in at least three model outputs are retained, ranked by vote count and evidence consistency, and truncated to at most five answers following the BioASQ factoid format.

\begin{table}[!htbp]
\caption{Evaluation on the BioASQ-13b test set for factoid questions. The best score in each column is highlighted in bold.}
\label{tab:13b_factoid}
\centering
\scriptsize
\resizebox{0.8\linewidth}{!}{%
\begin{tabular}{llrrr}
\toprule
Model & Method & Strict Acc. & Lenient Acc. & MRR \\
\midrule

\multirow{7}{*}{GPT-4o}
& Full snippets & 0.4230 & 0.4620 & 0.4420 \\
& Full snippets + ICL2 & \textbf{0.5380} & 0.5770 & \textbf{0.5580} \\
& Verifier + Reviser + ICL2 & \textbf{0.5380} & 0.5770 & \textbf{0.5580} \\
& Snippet highlight & 0.4620 & 0.5000 & 0.4810 \\
& Snippet highlight + ICL2 & \textbf{0.5380} & 0.5770 & \textbf{0.5580} \\
& Previous baseline & 0.4231 & 0.5000 & 0.4615 \\
& Previous baseline + self-reflection & 0.5000 & 0.5000 & 0.5000 \\

\midrule

\multirow{4}{*}{GPT-5.2}
& Full snippets & 0.4620 & \textbf{0.6150} & 0.5320 \\
& Full snippets + CoT ICL2 & 0.3850 & 0.4620 & 0.4230 \\
& Snippet filtering top-10 & 0.4620 & 0.5770 & 0.5060 \\
& Agent & 0.3460 & 0.3460 & 0.3460 \\

\midrule

\multirow{3}{*}{GPT-5.4}
& Full snippets & 0.5000 & 0.5770 & 0.5290 \\
& Snippet filtering top-10 & 0.5000 & \textbf{0.6150} & 0.5510 \\
& Agent & 0.3850 & 0.3850 & 0.3850 \\

\midrule

\multirow{2}{*}{Claude}
& Sonnet-4.6 Agent & 0.4620 & 0.5000 & 0.4810 \\
& Opus-4.6 Agent & 0.4620 & 0.5000 & 0.4810 \\

\bottomrule
\end{tabular}}
\end{table}

\subsection{List questions}

List questions are challenging because they require both high recall and high precision: the system must retrieve all valid entities while avoiding unsupported ones. As shown in Table~\ref{tab:13b_list}, agent-based variants achieved the strongest performance on the BioASQ-13b development benchmark, with GPT-5.2 obtaining the best mean precision, recall, and F-measure. These results motivated a collaborative list-agent pipeline for BioASQ-14b.

\begin{figure}[t]\centering\includegraphics[width=0.9\linewidth]{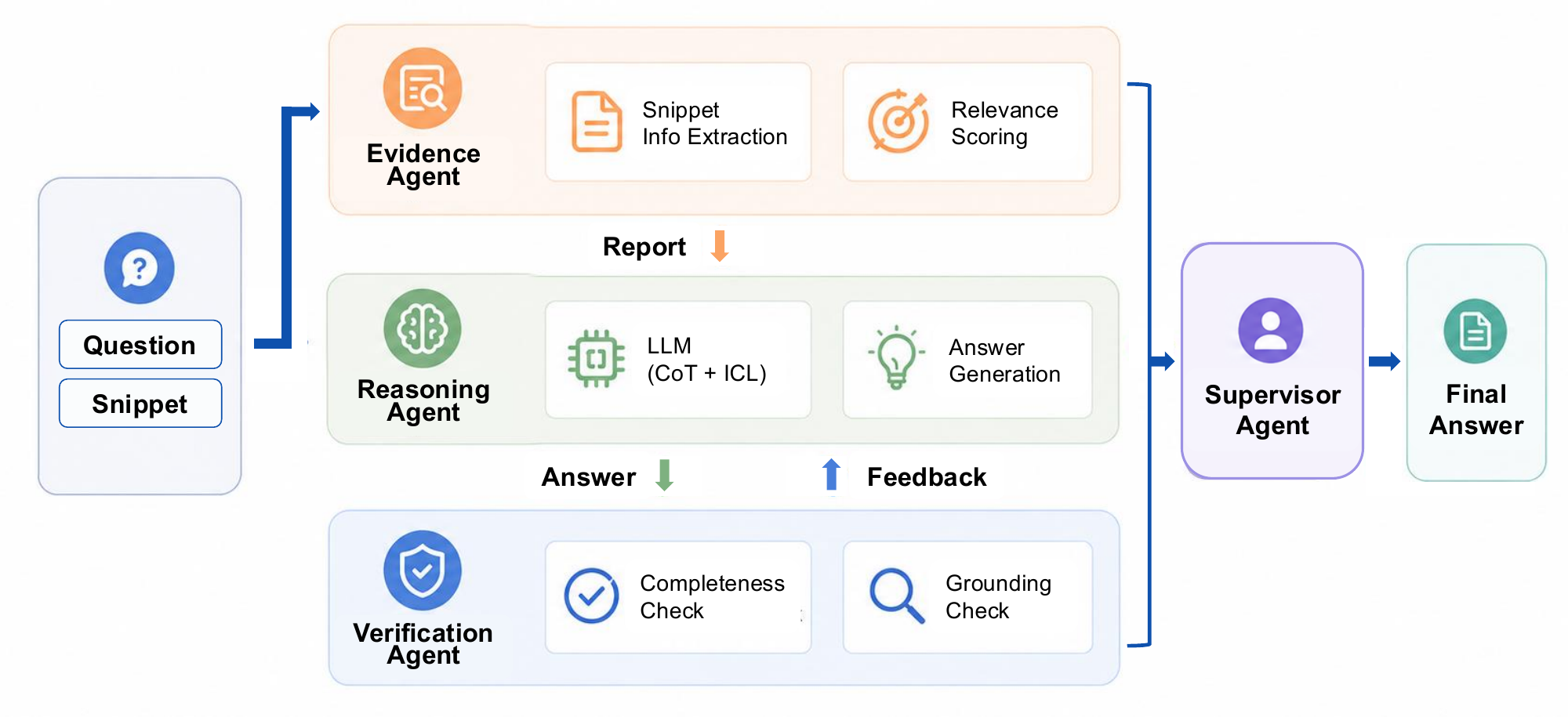}\caption{List-agent workflow for BioASQ list questions, combining Evidence Agent, Reasoning Agent, and Verification Agent. Verified outputs are reviewed by a supervisor to produce a final evidence-supported list answer.}\label{fig:list-results}\end{figure}

Figure~\ref{fig:list-results} illustrates the workflow. The pipeline consists of four sequential agents: an Evidence Analyst Agent, a Reasoning Agent, a Verification Agent, and a Supervisor Agent. 

The Evidence Analyst first scans all provided snippets, scores their relevance, extracts candidate entities, and produces a structured evidence report without committing to a final answer. The Reasoning Agent then uses this report, together with the original snippets, to generate a candidate list while independently re-reading the evidence to avoid omissions. The candidate list is passed to the Verification Agent, which audits it for grounding, completeness, formatting, and duplicate entities. Finally, the Supervisor Agent reviews the intermediate reports and verification feedback, then either adopts the candidate list or revises it using the original snippets. Representative prompt templates for these agents are provided in Appendix~\ref{app:prompt_examples}.

This separation of roles targets two common failure modes of single-pass LLM inference on list questions. Models may over-generate biomedical entities that are topically related but not directly requested, or under-generate by missing valid entities that appear only in low-salience snippets. By separating evidence extraction, answer generation, and entity-level verification, the list subsystem aims to improve recall and precision simultaneously.

\begin{table}[!htbp]
\caption{Evaluation on the BioASQ-13b test set for list questions. The best score in each column is highlighted in bold.}
\label{tab:13b_list}
\centering
\scriptsize
\resizebox{0.8\linewidth}{!}{%
\begin{tabular}{llrrr}
\toprule
Model & Method & Mean Prec. & Recall & F-measure \\
\midrule

\multirow{7}{*}{GPT-4o}
& Full snippets & 0.5377 & 0.5018 & 0.4923 \\
& Full snippets + ICL2 & 0.4938 & 0.4919 & 0.4613 \\
& Verifier + Reviser + ICL2 & 0.5229 & 0.4812 & 0.4765 \\
& Snippet highlight & 0.5487 & 0.4995 & 0.4975 \\
& Previous baseline & 0.5363 & 0.4947 & 0.4920 \\
& Previous baseline + self-reflection & 0.5000 & 0.4760 & 0.4643 \\

\midrule

\multirow{4}{*}{GPT-5.2}
& Full snippets & 0.4361 & 0.4096 & 0.4076 \\
& Full snippets + CoT ICL2 & 0.5768 & 0.5495 & 0.5398 \\
& Snippet filtering top-10 & 0.4601 & 0.4282 & 0.4289 \\
& Agent & \textbf{0.5918} & \textbf{0.5854} & \textbf{0.5560} \\

\midrule

\multirow{3}{*}{GPT-5.4}
& Full snippets & 0.4597 & 0.4463 & 0.4375 \\
& Snippet filtering top-10 & 0.4458 & 0.4384 & 0.4275 \\
& Agent & 0.5226 & 0.5730 & 0.5221 \\

\midrule

\multirow{2}{*}{Claude}
& Sonnet-4.6 Agent & 0.5023 & 0.5829 & 0.5117 \\
& Opus-4.6 Agent & 0.5241 & 0.5655 & 0.5185 \\

\bottomrule
\end{tabular}}
\end{table}

\subsection{Submission variants}

Across batches, submission variants were used to test small changes in prompting, answer normalization, and aggregation thresholds. For example, list-question variants differ in how strongly the verification feedback constrains the final list and in the precision-recall balance of the aggregation agent. Factoid variants differ mainly in in-context example selection and consensus thresholds. The official result page reports these variants separately as \texttt{ku\_dmis1}, \texttt{ku\_dmis2}, \texttt{ku\_dmis3}, \texttt{ku\_dmis4}, and \texttt{ku\_dmis5} when submitted.

\section{Results}
\subsection{Overall Performance}

\begin{table}[!htbp]
\caption{Best exact-answer scores on BioASQ-14b by batch, computed by taking the maximum score across submissions for each metric. ``Avg.'' denotes the arithmetic mean across the four batches.
}
\label{tab:best_scores}
\centering
\scriptsize
\resizebox{\textwidth}{!}{%
\begin{tabular}{lrrrrrrrrrr}
\toprule
\multirow{2}{*}{Batch} & \multicolumn{4}{c}{Yes/No} & \multicolumn{3}{c}{Factoid} & \multicolumn{3}{c}{List} \\
\cmidrule(lr){2-5}\cmidrule(lr){6-8}\cmidrule(lr){9-11}
 & Acc. & F1 Yes & F1 No & Macro F1 & Strict & Lenient & MRR & Mean Prec. & Recall & F-measure \\
\midrule
1 & 0.9412 & 0.9524 & \textbf{0.9231} & 0.9377 & 0.4348 & 0.4783 & 0.4565 & 0.3096 & 0.2618 & 0.2734 \\
2 & \textbf{0.9524} & \textbf{0.9655} & \textbf{0.9231} & \textbf{0.9443} & 0.2500 & 0.5000 & 0.3475 & 0.4672 & 0.4672 & 0.4628 \\
3 & 0.9091 & 0.9333 & 0.8571 & 0.8952 & 0.4118 & 0.5294 & 0.4706 & 0.4750 & 0.5535 & 0.4743 \\
4 & 0.8750 & 0.9000 & 0.8333 & 0.8667 & \textbf{0.4545} & \textbf{0.7273} & \textbf{0.5758} & \textbf{0.6677} & \textbf{0.7062} & \textbf{0.5652} \\
\midrule
Avg. & 0.9194 & 0.9378 & 0.8842 & 0.9110 & 0.3878 & 0.5588 & 0.4626 & 0.4799 & 0.4972 & 0.4439 \\
\bottomrule
\end{tabular}}
\end{table}

Table~\ref{tab:best_scores} summarizes the best exact-answer performance on BioASQ-14b across the four official batches. For each metric, the table reports the maximum score obtained among the submitted variants, with the final row showing the arithmetic mean across batches.

Overall, the yes/no subsystem produced the most stable results. Performance remained high across all batches, indicating that the snippet-shuffling ensemble and selective verification strategy were effective for binary decision making even under varying evidence conditions. Although later batches were more challenging, the system maintained strong macro-level performance, suggesting that it handled both positive and negative answers without relying heavily on one class.

Factoid performance showed greater variation across batches. The difference between strict and lenient accuracy suggests that the system often recovered the correct entity but did not always rank the exact expected answer first. This pattern reflects the difficulty of BioASQ factoid questions, where success depends not only on identifying the relevant biomedical entity but also on selecting the appropriate surface form and ranking it correctly. The strongest factoid results were observed in the final batch, suggesting that the combination of full-snippet reasoning, retrieved demonstrations, and multi-model consensus was particularly effective when answer candidates could be recovered from the provided evidence.

List questions exhibited the clearest improvement across batches. Both recall-oriented and precision-oriented metrics increased in later batches, with the final batch showing the strongest overall list performance. This trend supports the motivation behind the collaborative list-agent pipeline: separating evidence extraction, answer generation, verification, and supervision helps reduce both over-generation of unsupported entities and under-generation of valid but less salient entities.

\subsection{Variant Analysis}
\begin{table}[!htbp]
\caption{Submission variants that produced the best values in selected metric groups. Ties are shown when multiple variants obtained the same value.}
\label{tab:variant_sources}
\centering
\small
\begin{tabularx}{\textwidth}{llllll}
\toprule
Batch & Best yes/no metrics & Best factoid strict & Best factoid MRR & Best list precision & Best list F-measure \\
\midrule
1 & \texttt{ku\_dmis1} & \texttt{ku\_dmis1} & \texttt{ku\_dmis1} & \texttt{ku\_dmis1} & \texttt{ku\_dmis1} \\
2 & \texttt{ku\_dmis1} & all variants & all variants & \texttt{ku\_dmis1} & \texttt{ku\_dmis4} \\
3 & all variants & all variants & \texttt{ku\_dmis2--5} & \texttt{ku\_dmis1} & \texttt{ku\_dmis4,5} \\
4 & all variants & \texttt{ku\_dmis1,2,3,5} & \texttt{ku\_dmis3,5} & \texttt{ku\_dmis1} & \texttt{ku\_dmis4} \\
\bottomrule
\end{tabularx}
\end{table}

Table ~\ref{tab:variant_sources} provides a more detailed view of how different submission variants contributed to the final results across question types and metrics. For yes/no questions, the best-performing variant was largely unchanged across submissions: \texttt{ku\_dmis1} achieved the best scores in Batches 1 and 2, while all variants tied in Batches 3 and 4. This pattern suggests that the yes/no branch was relatively insensitive to the small prompt changes introduced in later submissions.

In contrast, factoid results show greater sensitivity to variant design, particularly for MRR. While strict accuracy was often shared by multiple variants, the best MRR in later batches came from more specific variants, such as \texttt{ku\_dmis2–5} in Batch 3 and \texttt{ku\_dmis3,5} in Batch 4. This indicates that different in-context example selections and consensus thresholds did not always change whether the correct entity appeared in the output, but they could affect its rank within the returned answer list. Since MRR rewards placing the correct entity higher, these differences suggest that factoid performance depends not only on entity detection but also on answer ordering and confidence calibration.
% In contrast, list-question performance depended more strongly on the submission variant. In several batches, the best precision and best F-measure were achieved by different variants. This reflects the inherent precision-recall trade-off in list answering: stricter verification can remove unsupported entities and improve precision, whereas more permissive aggregation can recover additional correct entities and improve recall.

List-question performance also depended on submission variants. The \texttt{ku\_dmis1} variant consistently produced the best precision, whereas \texttt{ku\_dmis4} or \texttt{ku\_dmis5} produced the best list F-measure in later batches. This reflects the intended precision–recall trade-off among list-question variants: a more conservative aggregation strategy can suppress unsupported entities and improve precision, while a less restrictive or more recall-oriented strategy can recover additional valid entities and improve F-measure. Batches 2 and 4 illustrate this clearly, where \texttt{ku\_dmis1} achieved the best precision, \texttt{ku\_dmis4} achieved the best F-measure.

Taken together, the results support the use of answer-type-aware inference strategies. The yes/no branch benefited from repeated inference and majority voting, the factoid branch benefited from full-snippet reasoning and multi-model consensus, and the list branch benefited from explicit evidence verification and aggregation. The remaining variability, especially in factoid and list questions, indicates that further improvements may require stronger biomedical synonym normalization, better ranking of factoid candidates, and more adaptive thresholding for list-answer aggregation.

\subsection{Question-type Analysis}

The results support the decision to use different inference strategies for different question types. The yes/no branch achieved high and stable macro F1 in the first two batches. The snippet-shuffling ensemble likely helped reduce ordering effects, while selective verification limited additional agentic processing to cases where the model pool disagreed. This design avoids applying a costly verification step to every question while still targeting uncertain samples.

Factoid performance was more variable. Batch 4 produced the highest MRR, but Batch 2 had lower strict accuracy and MRR. We attribute this behavior to two factors. First, factoid questions are highly sensitive to surface-form matching, so equivalent biomedical names may not be credited if the submitted form is not aligned with the official answer. Second, the consensus threshold of three model outputs improves precision but can remove rare correct answers mentioned by only a subset of models. The use of full snippets and chain-of-thought-based in-context learning helped stabilize entity extraction, but further improvements require stronger biomedical synonym normalization and abbreviation handling.

List-answer results improved substantially in later batches. The best F-measure increased from 0.2734 in Batch 1 to 0.5652 in Batch 4. This trend is consistent with the addition and refinement of the collaborative list-answer pipeline. The verification agent was particularly useful for removing unsupported or overly broad entities, while the aggregation agent allowed the system to recover valid entities that were initially omitted by the reasoning agent. The precision-recall trade-off across Batch 4 variants indicates that list answering remains sensitive to thresholding: the most precise variant was not the highest-recall variant, and the best F-measure came from a middle setting.

\section{Limitations}

This work has several limitations. First, the system relies on proprietary LLMs, so exact reproduction depends on access to the same models and versions. Second, the result summary in Table~\ref{tab:best_scores} is a per-metric maximum across submitted variants rather than the score of a single fixed system. This is useful for analyzing the best achieved metric values, but it should not be interpreted as one official system row. Third, the system focuses on exact answers and does not address ideal-answer generation; the official ideal-answer rows for \texttt{ku\_dmis} submissions were not scored in the result tables. Finally, because the method is prompt-driven, small prompt changes can affect output formatting, especially for factoid and list questions.

\section{Conclusion}

We presented the \texttt{ku\_dmis} system for BioASQ 14b Phase B exact-answer generation. The system uses question-type-specific LLM ensembles and verification mechanisms: snippet-shuffled majority voting for yes/no questions, full-snippet CoT-based in-context learning and consensus filtering for factoid questions, and a collaborative evidence-analysis and verification pipeline for list questions. Aggregating the official \texttt{ku\_dmis} submissions by taking the best value per metric, the system achieved average scores of 0.9110 yes/no macro F1, 0.4626 factoid MRR, and 0.4439 list F-measure across four batches. Future work should focus on reproducible open-model variants, biomedical synonym normalization, and unified optimization of precision-recall thresholds for list answers.

\section{Declaration on Generative AI}
During the preparation of this work, Generative AI tools were used for writing assistance, including grammar correction and sentence rephrasing. After using these tools, the authors reviewed and edited the content as needed and take full responsibility for the publication’s content.

%%
%% Define the bibliography file to be used
\bibliography{references}

%%
%% Prompt examples are moved out of the main methods section and placed at the end.
\appendix
\section{Prompt Examples}
\label{app:prompt_examples}
This appendix provides representative prompt templates used in the list-question agent pipeline. The templates are included for transparency; implementation details such as model-specific wrappers, retry logic, and JSON repair are omitted for brevity.

\begin{promptbox}{Evidence Analyst Agent Prompt}
{\bfseries\color{black!55}\normalsize~SYSTEM}\nobreak\\[2pt]
You are an Evidence Analyst specialized in biomedical literature. Your task is to carefully analyze provided snippets and extract key evidence relevant to answering a biomedical question. You do NOT answer the question itself --- you only analyze and organize the evidence.\\[6pt]
{\bfseries\color{black!55}\normalsize~USER}\nobreak\\[2pt]
Analyze the following snippets to extract evidence relevant to the question.\\[3pt]
{}[Question]: \{question\}\\
{}[Question Type]: list\\
{}[Snippet 1]: \dots\quad[Snippet 2]: \dots\quad\dots\\[3pt]
INSTRUCTIONS:\\
1. For each snippet, assess its relevance to the question (0.0 to 1.0).\\
2. Extract key entities, facts, and relationships from the relevant snippets.\\
3. Provide a concise evidence summary that organizes the extracted information.\\
4. Determine whether the evidence is sufficient to answer the question.\\
5. Extract ALL entities/items that belong to the category asked about. Scan every snippet exhaustively --- do not miss items.\\[3pt]
Respond ONLY in valid JSON with this exact format:\\
\{\\
\hspace*{1em}"relevant\_snippets": [\{"idx": 1, "relevance\_score": 0.95, "key\_facts": ["fact1", "fact2"]\}, \dots],\\
\hspace*{1em}"extracted\_entities": ["entity1", "entity2", \dots],\\
\hspace*{1em}"evidence\_summary": "A concise paragraph summarizing the key evidence\dots",\\
\hspace*{1em}"evidence\_sufficient": true,\\
\hspace*{1em}"confidence": 0.85\\
\}
\end{promptbox}
\promptcap{Prompt for the Evidence Analyst, which scores snippet relevance and extracts candidate entities.}

\begin{promptbox}{Reasoning Agent Prompt}
{\bfseries\color{black!55}\normalsize~SYSTEM}\nobreak\\[2pt]
You are a Reasoning Agent for BioASQ Task B. You receive an evidence analysis report and the original snippets. Your task is to generate the most accurate answer based on the evidence. You must ground your answer ONLY in the provided snippets --- never invent facts. You also explain your reasoning process and indicate which snippets support your answer.\\[6pt]
{\bfseries\color{black!55}\normalsize~USER}\nobreak\\[2pt]
You are generating an answer for a BioASQ list question.\\[3pt]
{}[Evidence Analysis Report]:\\
{}- Evidence Summary: \{evidence\_summary\}\\
{}- Key Entities: \{extracted\_entities\}\\
{}- Evidence Sufficient: \{evidence\_sufficient\}\\
{}- Analyst Confidence: \{confidence\}\\[3pt]
{}[Snippets]:\\
\{snippets\_text\}\\[3pt]
(optional ICL block: "Here are some examples of how to answer: \dots Now answer the following question in the same format:")\\[3pt]
{}[Question]:\\
\{question\}\\[3pt]
Use the evidence analysis to guide your reasoning, but base your final answer ONLY on the original snippets above. The evidence report helps you focus on what's important, but the snippets are the ground truth.\\[3pt]
Return up to 7--8 items. Scan every snippet exhaustively and do not rephrase entity surface forms. Respond ONLY in valid JSON:\\
\{\\
\hspace*{1em}"answer": ["item1", "item2", "item3", \dots],\\
\hspace*{1em}"reasoning": "\dots",\\
\hspace*{1em}"supporting\_snippets": [1, 2, 3],\\
\hspace*{1em}"confidence": 0.9,\\
\hspace*{1em}"alternative\_answers": [\dots]\\
\}
\end{promptbox}
\promptcap{Prompt for the Reasoning Agent, which drafts a grounded candidate list from the evidence report.}

\begin{promptbox}{Verification Agent Prompt}
{\bfseries\color{black!55}\normalsize~SYSTEM}\nobreak\\[2pt]
You are a Verification Agent for BioASQ Task B. You independently verify whether an answer is correctly grounded in the provided snippets. You check for factual accuracy, completeness, proper formatting, and consistency with the evidence analysis. You do NOT generate answers --- you only verify and provide feedback.\\[6pt]
{\bfseries\color{black!55}\normalsize~USER}\nobreak\\[2pt]
Verify the following answer for a BioASQ list question.\\[3pt]
{}[Question]: \{question\}\\[3pt]
{}[Snippets]:\\
\{snippets\_text\}\\[3pt]
{}[Answer to Verify]: \{answer\}\\[3pt]
{}[Reasoning Provided]: \{reasoning\}\\[3pt]
{}[Evidence Analysis Summary]: \{evidence\_summary\}\\
{}[Evidence Entities]: \{extracted\_entities\}\\[3pt]
VERIFICATION CHECKS:\\
1. GROUNDING: Is every claim in the answer directly supported by at least one snippet?\\
2. ACCURACY: Is the answer factually correct according to the snippets?\\
3. COMPLETENESS: Does the answer address the question fully?\\
4. FORMAT: Is the answer in the correct format for a list question?\\
5. CONSISTENCY: Is the answer consistent with the evidence analysis report?\\
6. LIST-SPECIFIC: Check completeness of the returned entities, flag any false positives, and remove duplicate entities.\\[3pt]
CROSS-CHECK: Compare the answer's entities with the Evidence Analyst's extracted entities. Flag any entities in the answer that don't appear in either the snippets or the evidence report.\\[3pt]
Respond ONLY in valid JSON:\\
\{\\
\hspace*{1em}"verdict": "supported" or "needs\_revision" or "unsupported",\\
\hspace*{1em}"issues": ["issue1", "issue2"],\\
\hspace*{1em}"suggested\_correction": "corrected answer or null if no correction needed",\\
\hspace*{1em}"confidence": 0.85\\
\}\\[3pt]
RULES:\\
{}- "supported": Answer is correct, grounded, and well-formatted.\\
{}- "needs\_revision": Answer has minor issues (format, verbosity, slight inaccuracy) but is mostly correct.\\
{}- "unsupported": Answer is factually wrong, contains hallucination, or completely misses the point.\\
{}- If verdict is "supported", issues should be empty and suggested\_correction should be null.\\
{}- If verdict is "needs\_revision" or "unsupported", provide specific issues and a suggested\_correction.
\end{promptbox}
\promptcap{Prompt for the Verification Agent, which audits the list for grounding, completeness, and duplicates.}

\begin{promptbox}{Supervisor Agent Prompt}
{\bfseries\color{black!55}\normalsize~SYSTEM}\nobreak\\[2pt]
You are the Supervisor Agent making the final decision for a BioASQ Task B question. You receive reports from three subordinate agents: (1) Evidence Analyst --- who analyzed the snippets, (2) Reasoning Agent --- who generated an answer, (3) Verification Agent --- who verified the answer. Your job is to synthesize these reports and produce the FINAL answer. You are the ultimate authority --- you may override any subordinate if you have good reason.\\[6pt]
{\bfseries\color{black!55}\normalsize~USER}\nobreak\\[2pt]
You are making the final decision for a BioASQ list question.\\[3pt]
{}[Question]: \{question\}\\[3pt]
{}[Original Snippets]:\\
\{snippets\_text\}\\[3pt]
=== SUBORDINATE AGENT REPORTS ===\\[3pt]
{}[Evidence Analyst Report]:\\
{}- Evidence Summary: \{evidence\_summary\}\\
{}- Key Entities: \{extracted\_entities\}\\
{}- Evidence Sufficient: \{evidence\_sufficient\}\\
{}- Confidence: \{confidence\}\\[3pt]
{}[Reasoning Agent Report]:\\
{}- Answer: \{answer\}\\
{}- Reasoning: \{reasoning\}\\
{}- Supporting Snippets: \{supporting\_snippets\}\\
{}- Confidence: \{confidence\}\\
{}- Alternative Answers: \{alternative\_answers\}\\[3pt]
{}[Verification Agent Report]:\\
{}- Verdict: \{verdict\}\\
{}- Issues: \{issues\}\\
{}- Suggested Correction: \{suggested\_correction\}\\
{}- Confidence: \{confidence\}\\[3pt]
=== DECISION RULES ===\\
1. If Verification verdict is "supported" with high confidence ($\geq$ 0.7):\\
\hspace*{1em}$\rightarrow$ Use the Reasoning Agent's answer as-is.\\
2. If Verification verdict is "needs\_revision" and a suggested\_correction exists:\\
\hspace*{1em}$\rightarrow$ Evaluate whether the correction is better than the original. Use your judgment.\\
3. If Verification verdict is "unsupported":\\
\hspace*{1em}$\rightarrow$ Check the suggested\_correction. If valid, use it.\\
\hspace*{1em}$\rightarrow$ Otherwise, re-derive the answer from the snippets yourself.\\
4. CROSS-CHECK: Compare entities in the Reasoning answer against the Evidence Analyst's extracted\_entities. If there's a mismatch, investigate and resolve.\\
5. ALWAYS verify your final answer is grounded in the original snippets.\\
6. When in doubt, prefer the answer with higher snippet support.\\[3pt]
FINAL ANSWER FORMAT: Return the final answer as a JSON array of entity strings.\\[3pt]
Respond ONLY in valid JSON:\\
\{\\
\hspace*{1em}"final\_answer": ["entity1", "entity2", \dots],\\
\hspace*{1em}"decision\_reasoning": "Brief explanation of why you made this decision\dots",\\
\hspace*{1em}"source": "reasoning" or "verification\_correction" or "supervisor\_override",\\
\hspace*{1em}"revision\_needed": false\\
\}\\[3pt]
IMPORTANT:\\
{}- "revision\_needed" should ONLY be true if you believe even your own final answer is unreliable and a complete re-analysis from scratch might help.\\
{}- In most cases, you should be able to produce a good final answer from the existing reports.
\end{promptbox}
\promptcap{Prompt for the Supervisor Agent, which synthesizes all reports into the final list answer.}

\end{document}